\pdfoutput=1

\documentclass[11pt]{article}

\usepackage[final]{acl}

\usepackage{times}
\usepackage{latexsym}

\usepackage[T1]{fontenc}

\usepackage[utf8]{inputenc}

\usepackage{microtype}

\usepackage{inconsolata}
\usepackage{booktabs}
\usepackage{graphicx}
\title{AdaptEval: Evaluating Large Language Models on Domain Adaptation for Text Summarization}

\author{Anum Afzal \\
  Technical University of Munich \\
  \texttt{anum.afzal@tum.de} \\ 
  \And
  Ribin Chalumattu \\
  ETH Zürich \\
  \texttt{cribin@inf.ethz.ch} 
  \AND
  Florian Matthes \\
  Technical University of Munich \\
  \texttt{matthes@tum.de} \\ \\\And
  Laura Mascarell \\
  ETH Zürich \\
  \texttt{lmascarell@inf.ethz.ch} \\}

\begin{document}
\maketitle
\begin{abstract}
Despite the advances in the abstractive summarization task using Large Language Models (LLM), there is a lack of research that assess their abilities to easily adapt to different domains. We evaluate the domain adaptation abilities of a wide range of LLMs on the summarization task across various domains in both fine-tuning and in-context learning settings. We also present AdaptEval, the first domain adaptation evaluation suite. AdaptEval includes a domain benchmark and a set of metrics to facilitate the analysis of domain adaptation. Our results demonstrate that LLMs exhibit comparable performance in the in-context learning setting, regardless of their parameter scale. 
\end{abstract}

\section{Introduction}
Large Language Models (LLM) have achieved remarkable improvements on a wide range of natural language processing tasks, including abstractive text summarization, the task of generating an abridged version of the most relevant information in a document~ \cite{basyal2023text}. Recent works study the domain adaptation abilities of LLMs on the summarization task. However, the research is still limited to a single domain, such as news articles~\cite{goyal2022news,zhang2023benchmarking} or clinical reports~\cite{van2023clinical}. We argue that there is a lack of research across domains to better understand the abilities of these models to adapt to different targets.

In this paper, we assess the domain adaptation abilities of 11 models, including conventional encoder-decoder models and a wide range of LLMs in various parameter sizes, on the summarization task. In particular, we experiment with fine-tuning and in-context learning (ICL) settings and evaluate their performance across various domains (i.e.\ governmental, medical, and scientific), reporting scores on a collection of automatic\textemdash ROUGE~\cite{lin-2004-rouge} and BERTScore~\cite{zhang2019bertscore}\textemdash and domain adaptation metrics.~The latter includes domain vocabulary overlap~\cite{yu-etal-2021-adaptsum}, and our adaptations of G-eval~\cite{liu-etal-2023-g} and token distribution shift~\cite{lin2023unlocking} to the task.

The experimental results show the abilities of LLMs to adapt to the domain in the ICL setting. In particular, \textit{small} models with 7b parameters achieve comparable performance to their larger counterparts with only two learning examples. However, G-eval highlights the difficulty of adapting to the medical domain. While the fine-tuned models achieve the best performance in terms of automatic scores, their adaptation to the domain vocabulary is inferior to the ICL setting. ~Finally, we release the domain benchmark and evaluation metrics as the first domain \textbf{Adapt}ation \textbf{Eval}uation suite (\textbf{AdaptEval}) to facilitate the evaluation of models and foster further research on this task.\footnote{AdaptEval code is available on \href{https://github.com/anum94/AdaptEval}{AdaptEval}.} 

\section{The Domain Adaptation Suite}

\subsection{Domains Benchmark}
Our benchmark contains data from different datasets on the scientific, medical, and governmental domains. The final size of the domain datasets is listed in Table~\ref{tab:dataset_size}, after removing instances with extractive summaries, or extremely long summaries or sources as in \citet{shaham-etal-2022-scrolls}.\footnote{Deleted: 3\% arXiv, 4\% PubMed, and 0.4\% GovReport.}

\paragraph{Science} The data consists of scientific articles from the arXiv platform, where the human-written abstracts are used as reference summaries of the articles~\cite{cohan-etal-2018-discourse}.

\paragraph{Medical} The medical domain comprises academic articles in the field of biomedical and life sciences from the PubMed dataset~\citep{cohan-etal-2018-discourse}. Similarly to arXiv, the article abstracts are regarded as abstractive summaries.   

\paragraph{Government} The data comes from the GovReport dataset, a collection of reports on national policy issues paired with human-written executive summaries~\cite{huang-etal-2021-efficient}. The documents are 1.5 and 2.5 times longer than those from arXiv and PubMed, respectively.

\subsection{Evaluation Metrics}
The suite provides a set of metrics to evaluate the performance of summarization models and approaches across domains. Specifically, we include the standard summarization metrics ROUGE~\cite{lin-2004-rouge} and BERTScore~\cite{zhang2019bertscore}, which measure n-gram and contextual similarity against a reference, respectively. To get better insights into their domain adaptation abilities, we also implement several metrics that assess the domain language. We describe them in the rest of the section.


\paragraph{Domain Vocabulary Overlap (DVO)} We compute the percentage of domain vocabulary in the generated output as in \citet{yu-etal-2021-adaptsum}. The domain vocabulary consists of the top 10k most frequent words in the domain excluding stopwords.

\begin{table}
    \small
    \centering
    \begin{tabular}{lrrr}
    \toprule
    \textbf{Domain} & \multicolumn{1}{c}{Train} & \multicolumn{1}{c}{Val.} & \multicolumn{1}{c}{Test} \\
    \midrule
    Science  & 203,037  & 6,436 & 6,440 \\
    Medical &  119,924  & 6,633 & 6,658 \\
    Government &  17,517  & 973 & 973 \\
    \bottomrule
    \end{tabular}
    \caption{Sizes of domain datasets.}
    \label{tab:dataset_size}
\end{table}


\begin{table}
\small
    \centering
    \begin{tabular}{lrrr}
        \toprule
        \textbf{Domain} & \multicolumn{1}{c}{Size} & \multicolumn{1}{c}{\#W} & \multicolumn{1}{c}{\#Sum W}\\
        \midrule
        Science       & 215,913 & 6,029.9 & 272.7 \\
        Medical      & 133,215 & 3,049.9 & 204.4 \\
        Government   & 19,466 & 9,409.4 & 553.4 \\
        \bottomrule
    \end{tabular}
\caption{Total sizes of the domain datasets and average word count of source (\#W) and summary (\#Sum W).}
\label{tab:dataset_size_analysis}
\end{table}
\paragraph{Domain Token Distribution Shift} \citet{lin2023unlocking} analyzes the impact of LLM alignment and proposes to measure the token distribution shifts between base models and their aligned counterparts. We adopt the token distribution shift approach to domain adaptation. Specifically, we focus on the domain vocabulary (i.e.\ 10k most frequent words) and analyze the effects of adaptation strategies, such as ICL and fine-tuning on their distribution. 

Formally, given a prompt $p$, we first use the fine-tuned model to generate a summary by greedy decoding, where the summary is represented as a sequence of tokens $S = \{s_0,...,s_T\}$ from the model vocabulary $\mathcal{V}$, such that $s_t \in \mathcal{V}$ for $0 < t < T$. Next, we process each token in $S$ sequentially. At each step $t$, we get the probability distribution of the next token prediction given $p$ and the prior context $p(\cdot\ |\ \mathbf{s}_{<t},\mathbf{p})$ using both fine-tuned and base models. In the in-context learning setting, we use the same model, but the adapted approach extends the prompt $p$ with learning examples.\footnote{The method can also be applied to compare models of different parameter scales in different adaptation settings.} Finally, we rank the tokens in both distributions according to their probability and provide \textit{KL-divergence} scores and the \textit{token shift rate} of those tokens in the vocabulary domain. While the former represents their distribution similarity, the latter computes the frequency at which the adapted approach predicts a token from the vocabulary domain that is not among the top three predictions of the base model.


\paragraph{Reference-free evaluation with GPT-4} G-eval uses GPT-4~\cite{openai2023gpt4} with chain-of-thought prompting~\cite{NEURIPS2022_9d560961} to evaluate summaries across quality features, such as coherence or fluency, achieving high correlation with human judgments~\cite{liu-etal-2023-g}. Similarly, we design a prompt to score the degree to which a summary adheres to the domain language on a scale from 1 to 5. Our prompt includes the reasoning steps generated by GTP-4 as in~\citet{liu-etal-2023-g} (see Appendix~\ref{apx:prompt}).

\begin{table*}[hbt]
\small
\centering
\begin{tabular}{lcccccccccc}
\toprule
& \multicolumn{3}{c}{\textbf{Medical}} & \multicolumn{3}{c}{\textbf{Science}} & \multicolumn{3}{c}{\textbf{Government}}\\
\cmidrule(lr){2-4}\cmidrule(lr){5-7}\cmidrule(lr){8-10}
& BERTScore & DVO & ROUGE & BERTScore & DVO& ROUGE & BERTScore & DVO & ROUGE \\
\midrule
\multicolumn{10}{c}{\textit{Zero-shot Setting}} \\

\midrule

\texttt{PEGASUS-X} & {0.690} & {6.28} & {3.55}& {0.538}  & {11.98} & {5.85}& {0.736}  & {5.58} & {9.06}\\ 
\texttt{Falcon~~7b} & {0.811} & {31.87}  & {13.68}& {0.810}  & {30.16} & {14.54}& {0.821}  & {31.49} &           {13.86}\\
\texttt{Llama2~~7b} & {0.783} & {21.15}  & {10.94}& {0.818}  & {28.61} & {18.33}& {0.845}  & {34.36} & {18.86}\\
\texttt{Mistral 7b} & {0.788} & {24.78}  & {9.44}& {0.806}  & {28.81} & {13.68}& {0.815}  & {31.18} & {12.02}\\
\texttt{Vicuna~~7b} & {0.727} & {9.49}  & {2.11}& {0.781}  & {23.94} & {7.93}& {0.813}  & {30.69} & {10.80}\\
\texttt{Llama2 13b} & {0.764} & {20.78}  & {6.26}& {0.783}  & {23.48} & {8.58}& {0.797}  & {24.04} & {10.80}\\
\texttt{Vicuna 13b} & {0.745} & {15.76}  & {1.58}& {0.763}  & {19.07} & {4.43}& {0.783}  & {27.18} & {7.17}\\
\texttt{Falcon 40b} & {0.816} & {35.51}  & {13.85}& {0.822}  & {34.98} & {17.59}& {0.827}  & {35.51} &          {13.85}\\
\texttt{Llama2 70b} & \textit{0.842} & {35.50}  & \textit{24.59}& \textit{0.837}  & {35.22} & {23.35}& \textit{0.855}  & {36.05} & {21.48}\\
\texttt{ChatGPT} & \textbf{0.844} & {36.69}  & \textbf{24.81}& \textbf{0.838}  & {36.58} & \textbf{23.95}& \textbf{0.859}  & {37.73} & \textbf{22.34}\\
\texttt{GPT-4o mini} & {0.843} & \textbf{41.04} & {22.26}& {0.834}  & \textbf{40.85} & {20.16}& {0.856}  & \textbf{41.51} & {21.12}\\
\midrule
\multicolumn{10}{c}{\textit{Two-shot Setting}} \\
\midrule

\texttt{Llama2~~7b} & {0.819} & {35.95}  & {21.11}& {0.824}  & {35.34} & {20.92}& {0.847}  & {30.22} & {17.39}\\
\texttt{Mistral 7b} & \textit{0.816} & {32.05}  & {21.30}& {0.802}  & {23.61} & {17.76}& {0.844}  & {30.08} & \textbf{19.21}\\
\texttt{Vicuna~~7b} & {0.831} & {36.29}  & {21.54}& {0.827}  & {34.65} & {20.31}& {0.851}  & {30.28} & {17.29}\\
\texttt{Llama2 13b} & \textit{0.820} & {35.02}  & {19.00}& {0.809}  & {32.30} & {18.97}& {0.814}  & {29.92} & {14.30}\\
\texttt{Vicuna 13b} & {0.822} & {35.51}  & {19.69}& \textit{0.807}  & {33.32} & {14.86}& {0.789}  & {29.34} & {8.34}\\
\texttt{Llama2 70b} & \textbf{0.845} & 37.61  & \textit{22.40}& \textbf{0.842}  & {36.65} & {23.03}& {0.851}  & {29.59} & {18.72}\\
\texttt{ChatGPT}  & {0.841} & \textbf{38.58}  & {22.92}& {0.837}  & \textbf{38.39} & \textbf{23.15}&  \textbf{0.853}  & \textbf{30.44} & {16.82}\\
\texttt{GPT-4o mini} & {0.842} & {30.64} & \textbf{23.18}& {0.835}  & {29.14} & {21.47}& {0.850}  & {30.40} & {16.04}\\

\midrule
\multicolumn{10}{c}{\textit{Fine-tuning Setting}} \\
\midrule
\texttt{BART} & {0.852} & \textbf{37.03} & {24.80}& {0.844}  & \textit{34.15} & {22.20}& {0.856}  & {25.14} & {28.44}\\ 
\texttt{PEGASUS-X} & {0.850} & {28.72} & \textbf{31.18}& {0.852}  & \textbf{34.61} & \textbf{28.11}& \textbf{0.868}  & {22.07} & \textbf{31.98}\\ 
\texttt{Llama2~~7b$^{1}$}& {0.859} & {33.61}  & {25.81}& {0.858}  & {33.06} & {25.30}& {0.850}  & {29.30} & {24.81}\\
\texttt{Llama2~~7b$^{2}$}& {0.861} & {35.15}  & {26.00}& {0.856}  & {30.49} & {25.46}& {x}  & {x} & {x}\\
\texttt{Llama2~~7b$^{3}$}& \textit{0.862} & {33.71}  & {26.81}& {0.854}  & {27.43} & {25.35}& {x}  & {x} & {x}\\
\texttt{Mistral 7b$^{2}$} & \textbf{0.863} & {35.81}  & {27.17}& \textbf{0.863}  & {34.00} & {27.29}& {0.833}  & {21.66} & {23.08}\\
\texttt{Llama2 13b$^{2}$}& \textit{0.862} & \textit{35.28}  & {26.26}& \textit{0.860}  & {32.67} & {26.47}& {x}  & {x} & {x}\\

\bottomrule
\end{tabular}
\caption{BERTScore F$_1$, DVO (\%), and the geometric mean of ROUGE-1/2/L (ROUGE) of all models across the three domains. The value `x' implies that the model was not evaluated under those settings. $^{1}$/$^{2}$/$^{3}$ indicate fine-tuning with 1k, 5k, and 10k instances, respectively.}
\label{tab:evaluation scores}
\end{table*}

\section{Domain Adaptation Task}
We assess the performance of 11 models across domains in both fine-tuning\footnote{We exclude GovReport from fine-tuning on 5k and 10k samples, since the train set doesn't have enough documents to fit into the models context window of 4096 tokens\textemdash only 1148 instances with maximum 4k length in the training split.} and ICL settings.

\subsection{Models Selection}
We select a wide variety of models from the conventional encoder-decoder transformer models\textemdash BART~\cite{lewis-etal-2020-bart} and PEGASUS-X~\cite{phang2022investigating}\textemdash to the recent instruction-based LLMs. The latter includes open-source models from the Llama2 family~\cite{touvron2023llama}, Vicuna~\cite{vicuna2023}, Falcon~\cite{almazrouei2023falcon}, and Mistral AI~\cite{jiang2023mistral}. For each model family, we consider various model sizes ranging from 7b to 70b parameters, if available. Additionally, we consider the close-source model ChatGPT from OpenAI. We provide the checkpoints and technical details in Appendix~\ref{sec:appendix-training-details}.


        

\begin{table*}[h]
    \centering
    \small
    \begin{tabular}{llcccccc}
        \toprule
        &  & \multicolumn{2}{c}{\textbf{Science}} & \multicolumn{2}{c}{\textbf{Medical}} & \multicolumn{2}{c}{\textbf{Government}} \\
        \cmidrule(lr){3-4}\cmidrule(lr){5-6}\cmidrule(lr){7-8}
        \multicolumn{1}{c}{\textit{base}} & \multicolumn{1}{c}{\textit{2-shot}} & KL & TSR & KL & TSR & KL & TSR\\
        \midrule
        \texttt{Llama2~~7b} & vs. \texttt{7b}  & {19.70}& {92.14}& {19.27}& {97.44} & {17.40} & {94.33} \\
        \texttt{Mistral 7b} & vs. \texttt{7b} &{13.88} &{91.33} &{14.01} &{95.40} & {13.40} & {90.00} \\
        \texttt{Vicuna~~7b} & vs. \texttt{7b}  &{17.67} &{92.35} &{18.32} & {93.89} & {15.42} & {94.04} \\
        \texttt{Llama2 13b} & vs. \texttt{13b} &{15.58} &{96.95} & {16.53} & {96.76} & {14.67} & {98.82} \\
        \texttt{Vicuna 13b} & vs. \texttt{13b}  & {18.12} & {97.13} &{17.34} & {90.70} & {16.79} & {99.10} \\
        \texttt{Llama2 70b} & vs. \texttt{70b}  & {16.78}& {95.68} & {17.12} &  {98.19}& {13.10} & {92.36} \\    

        \midrule
        \multicolumn{1}{c}{\textit{2-shot}} & \multicolumn{1}{c}{\textit{2-shot}} & KL & TSR & KL & TSR & KL & TSR\\
        \midrule
        \texttt{Llama2 13b} & vs. \texttt{7b} & 0.21 & 2.87 & 0.38 & 1.67 & {0.32} & {10.38} \\
        \texttt{Vicuna 13b} & vs. \texttt{7b} &  0.25& 2.07& 0.38 & 4.57 & {0.24} & {0.00} \\
        \texttt{Llama2 70b} & vs. \texttt{13b}  & 0.47 & 5.18 & 0.31 & 3.50 & {0.49} & {4.92} \\
        \texttt{Llama2 70b} & vs. \texttt{7b} & 0.43 & 3.92 & 0.46 &5.01 & {0.54} & {6.88} \\

        \midrule
        \multicolumn{1}{c}{\textit{base}} & \multicolumn{1}{c}{\textit{FT}} & KL & TSR & KL & TSR & KL & TSR  \\
        \midrule
        \texttt{Llama2~~7b} & vs. \texttt{7b} & 0.81 & 12.40  & 0.35 & 4.70 & {21.49}& {15.15} \\
        \texttt{Mistral 7b} &  vs. \texttt{7b} & 0.52 & 11.54 & 0.37 & 4.42& {0.18}& {3.21} \\
        \texttt{Llama2 13b} & vs. \texttt{13b} & 0.51 & 6.84  & 0.48& 7.32& {x}& {x} \\
        \bottomrule
    \end{tabular}
    \caption{Effect of different model sizes, two-shot in-context learning, and Fine-Tuning in terms of token distribution shift scores\textemdash KL divergence and Token Shift Rate (\%) calculated over 10 samples. Two-shot has the major impact on the models' predictions. The low scores between different model sizes indicate that parameter size does not have a significant effect on domain adaptation in the two-shot setting. }
    \label{tab:tds-scores}
    \end{table*}

\subsection{Results}
\label{sec:results}
Table~\ref{tab:evaluation scores} shows the performance of the models across domains in terms of ROUGE, BertScores, and DVO. We observe that the model size has a direct impact on their overall performance in the zero-shot setting; however, this performance gap is considerable reduced in the ICL setting with only two learning examples. In fact, the scores of the small 7b models are comparable to the large Llama 70b or the even larger ChatGPT. To validate these results, we compute the token distribution shift between models of different sizes in the two-shot setting (Table~\ref{tab:tds-scores}). The scores reflect that their probability distributions are very similar, confirming that there are no major differences in their performance.

In contrast, the fine-tuning results in Table~\ref{tab:evaluation scores} are mixed. Overall, the models outperform their counterparts in the two-shot setting in terms of ROUGE scores; however, there is a decrease in DVO. In particular, PEGASUS-X achieves the best ROUGE scores. We argue that this is attributed to the model's fine-tuning process, since the parameters are adjusted to optimize on ROUGE. Additionally, BART achieves the highest DVO despite its small parameter size (110M). \citet{johner-etal-2021-error} point out to the model's tendency to generate highly extractive summaries, which favours the use of domain vocabulary. Finally, the token shift rate and KL-divergence scores between the base and fine-tuned models are higher than in the two-shot setting. However, we observe that most distribution shifts are due to stylistic tokens, as also reported in \citet{lin2023unlocking} between the base and their aligned LLMs.

To confirm these findings, we also evaluate the summaries using GPT-4 shown in Table~\ref{tab:g-eval}, which have a strong correlation with human judgments, along with our addition to measure domain adaptation, on a random sample of 25 articles.\footnote{Due to the costs of using GPT-4 with large prompts, we only report the scores on four models outputs of 25 random instances.} The scores on arXiv data are consistent with our previous results, showing that ICL achieves the best performance, and the model parameter size does not have a significant impact. However, PubMed obtains remarkably low scores, which highlights the difficulty of the models to adapt to the medical domain. The LLMs however, find it easier to adapt to the Government domain.



\begin{table*}[h]
\resizebox{\textwidth}{!}{%
\centering
    \begin{tabular}{lccc ccc ccc}
    \toprule
     & \multicolumn{3}{c}{\textbf{DA (ours)}} & \multicolumn{3}{c}{\textbf{Coherence}} & \multicolumn{3}{c}{\textbf{Fluency}} \\ 
     \cmidrule(lr){2-4}\cmidrule(lr){5-7}\cmidrule(lr){8-10}
    \multicolumn{1}{c}{\textit{2-shot}} & arXiv & PubMed & GovReport & arXiv & PubMed & GovReport & arXiv & PubMed &GovReport\\
    \midrule
    \texttt{Llama2~~7b} & 4.20 & 1.0 & 4.04 & 3.80 & 2.0 & 3.96 & 2.72 & 2.0 & 2.96\\
    \texttt{Llama2 70b} & 3.96 & 1.0 & 4.40 & 3.20 & 1.0  & 3.96& 2.56 &1.0 & 3.00\\
    \midrule
    \multicolumn{1}{c}{\textit{FT}} & & & & & &  & & & \\
    \midrule
    \texttt{Llama2~~7b} & 3.48 & 2.0  & 4.16 & 2.08 & 2.0  & 3.40 & 2.04 & 2.0  & 2.84\\
    \texttt{PEGASUS-X} & 3.88 & 2.8  & 4.40 & 2.88 & 2.0 & 3.72 & 2.40 & 2.0 &2.72\\
    \bottomrule
    \end{tabular}
    }
    \caption{Evaluation scores using GPT-4 on 25 random samples from the arXiv, PubMed and GovReport datasets in terms of coherence (1-5), fluency (1-3), and our Domain Adaptation (DA) (1-5). }
    \label{tab:g-eval}
 
\end{table*}

\subsection{Manual Evaluation}
Two in-house domain experts perform a blind manual evaluation of the same arXiv samples used in GPT-4 evaluation (Table~\ref{tab:g-eval}). The setting comprises of 25 random arXiv articles paired with four different summaries generated with Llama2 (7b and 70b) in the two-shot setting, fine-tuned Llama2 (7b) and PEGASUS-X. To avoid biases, we randomly shuffle the evaluation instances and their summaries for each annotator. 

We ask the annotators to rank the generated summaries according to how well the vocabulary and style of the outputs adapt to the scientific domain. The task is especially challenging when the summaries contain similar vocabulary. Therefore, we focus on the relative performance of the models; that is, their agreement on an output being ranked higher than the other. The final Cohen's $\kappa$ inter-annotator agreement is 0.4. The results show that the annotators consistently rated the outputs of both Llama2 7b and 70b in the two-shot scenario among the top two positions of the ranking\textemdash 60\% and 52\%, respectively\textemdash whereas the fine-tuned models were the least preferred\textemdash only 12\% (Llama2 7b) and 16\% (Pegasus-X) rated on top.

\section{Related Work}
Some recent works evaluate the domain adaptation abilities of LLMs on the summarization task, albeit limited to a specific domain. \citet{van2023clinical} focus on clinical data and tackle the summarization of electronic health records. They evaluate eight different LLMs across six datasets in the same domain. \citet{fu2024tiny} investigate whether model size has an impact on the summarization performance of business meeting
transcripts. The results show that smaller LLMs cannot outperform their larger counterparts (from 7b to 70b parameters), even after fine-tuning, except for FLAN-T5 with 780M parameters~\cite{chung2022scaling}. In contrast, \citet{zhang2023benchmarking} provides a benchmark for text summarization of news articles and concludes that instruct-tuning rather than model size is the key to text summarization with LLMs. Similarly, \citet{goyal2022news} propose also a news summarization benchmark and compare the performance between conventional encoder-decoder and instruction-based models. Prior to the LLM era, \citet{yu-etal-2021-adaptsum} explored domain adaptation techniques in a low-resource setting, such as fine-tuning and second pre-training of encoder-decoder summarization models on a wide range of datasets. 
\section{Conclusion}
We evaluate the domain adaptation abilities of Large Language Models across scientific, medical, and governmental domains using a set of adapted evaluation metrics. Additionally, we release AdaptEval, an evaluation suite that facilitates the analysis of domain adaptation. Our experiments show that smaller LLMs exhibit domain-shift challenges, but they are able to achieve comparable performance to larger LLMs when provided with only two learning examples. In contrast, fine-tuning does not have a significant impact on the vocabulary domain, but only on stylistic tokens. Overall, the G-eval scores indicate that the medical domain is challenging for these models. We expect our work to encourage and facilitate further research on domain adaptation with LLMs across domains. We plan to continue this research in future work. 


\section*{Limitations}
To fairly compare the performance of the different models, we generally restricted our evaluation to those models with context window of 4096. An exception is the language model BART with a context window of 1024. 
Additionally, due to the high costs of performing human evaluations on multiple domains, we only annotated ArXiv data to reaffirm the results obtained through the automatic metrics. Our goal is to facilitate the evaluation of models across domains to the research community. Therefore, our suite consists of a set of metrics to evaluate domain adaptation and general summarization quality, allowing for a comprehensive comparison of the models performance on multiple datasets. Lastly, given the cost associated with GPT-4, we performed LLM-based evaluation on only 25 random samples.

\section*{Ethics Statement}
Throughout our experiments, we strictly adhere to the ACL Code of Ethics. Since we used already established open-source benchmark datasets, the concern of privacy does not apply. The manual evaluation was performed by in-house domain experts, who receive a full salary. They were informed about the task and usability of data in the research. Their annotations were stored anonymously, mitigating any privacy concerns. Through our fine-tuning strategies, no additional bias was introduced into the models, other than what might already be part of the model weights or the benchmark dataset. The goal of the research is to evaluate the domain adaptation capabilities of existing models on a text summarization task. The results and discussions in this paper are meant to further promote research in the area of domain-specific language modeling with an over-arching goal of bridging the gap between academia and application. All training scripts and trained models will be made available to the research community.


\section*{Acknowledgements}
This project is supported by Ringier, TX Group, NZZ, SRG, VSM, viscom, the ETH Zurich Foundation, and the German Federal Ministry of Education and Research (BMBF) grant 01IS17049 Software Campus 2.0 (TU München). 

\bibliography{anthology,custom}
\newpage
\appendix

\section{Technical Details}
\label{sec:appendix-training-details}
The fine-tuning and inference procedure was done by leveraging Nvidia A100-80GB GPUs.

\subsection{Zero-shot Setting}
We used the instruct-tuned or chat versions of the models. As for ChatGPT, we used the OpenAI API\footnote{\url{https://platform.openai.com/}} and the latest snapshot available, \texttt{gpt-3.5-turbo-0613} from June 13th, 2023. For zero-shot setting, we used Llama2 (7b)\footnote{\url{https://huggingface.co/meta-llama/Llama-2-7b-chat-hf/}}, Llama2 (13b)\footnote{\url{https://huggingface.co/meta-llama/Llama-2-13b-chat-hf/}}, Llama2 (70b) \footnote{\url{https://huggingface.co/meta-llama/Llama-2-70b-chat-hf/}}, Vicuna (7b)\footnote{\url{https://huggingface.co/lmsys/vicuna-7b-v1.5}}, Vicuna (13b)\footnote{\url{https://huggingface.co/lmsys/vicuna-13b-v1.5}}, Falcon (7b)\footnote{\url{https://huggingface.co/tiiuae/falcon-7b}}, Falcon (40b)\footnote{\url{https://huggingface.co/tiiuae/falcon-40b}}, and Mistral AI (7b)\footnote{\url{https://mistralai/Mistral-7B-Instruct-v0.1}}.

When generating summaries, we sample a maximum of 256 tokens for the arXiv and PubMed datasets, while scaling to 1024 tokens for the GovReport dataset, as is standard procedure in other contemporary publications. The prompts used 0-shot and 2-shot settings for generating the summaries is shown in Table~\ref{tab:prompt}.

\subsection{In-context Learning Setting}
We used the same model checkpoints as the ones from zero-shot settings for in-context learning. We excluded Falcon from in-context learning, since its context window of 2048 is too small to fit 2 learning examples.

\subsection{Fine-tuning Setting}
The links to all fine-tuned models is displayed in \autoref{tab:finetuned-model-links}.

\begin{table*}[h]
    \centering
    \small
    
    \begin{tabular}{lccc}
        \toprule
        & {\textbf{Science}} & {\textbf{Medical}} & {\textbf{Government}} \\
        \midrule
        \texttt{BART}  & \href{https://huggingface.co/mtc/bart-base-arxiv-1024}{bart-arxiv-1024} & \href{https://huggingface.co/mtc/bart-base-pubmed-1024}{bart-pubmed-1024} & \href{https://huggingface.co/mtc/bart-base-govreport-1024}{bart-govreport-1024} \\
        
        \texttt{PEGASUS-X}  & \href{https://huggingface.co/twigs/bigbird-pegasus-large-4096-arxiv}{bigbird-pegasus-arxiv-4096} & \href{https://huggingface.co/twigs/bigbird-pegasus-large-4096-pubmed}{bigbird-pegasus-pubmed-4096} & \href{https://huggingface.co/twigs/bigbird-pegasus-large-4096-govreport}{bigbird-pegasus-govreport-4096} \\
        
        \texttt{Llama2~~7b$^{1}$} & \href{https://huggingface.co/mtc/meta-llama-Llama-2-7b-hf-arxiv-summarization-1000-last_merged}{Llama-2-7b-arxiv-4096} & \href{https://huggingface.co/mtc/meta-llama-Llama-2-7b-hf-pubmed-summarization-1000-last_merged}{Llama-2-7b-pubmed-4096} & \href{https://huggingface.co/mtc/meta-llama-Llama-2-7b-hf-govreport-summarization-1000-last_merged}{Llama-2-7b-govreport-4096} \\ 
        
        \texttt{Llama2~~7b$^{2}$} & \href{https://huggingface.co/mtc/meta-llama-Llama-2-7b-hf-arxiv-summarization-5000-last_merged}{Llama-2-7b-arxiv-4096} & \href{https://huggingface.co/mtc/meta-llama-Llama-2-7b-hf-pubmed-summarization-5000-last_merged}{Llama-2-7b-pubmed-4096} & x\\ 
        
        \texttt{Llama2~~7b$^{3}$} & \href{https://huggingface.co/mtc/meta-llama-Llama-2-7b-hf-arxiv-summarization-10k-last_merged}{Llama-2-7b-arxiv-4096} & \href{https://huggingface.co/mtc/meta-llama-Llama-2-7b-hf-pubmed-summarization-10k-last_merged}{Llama-2-7b-hf-pubmed-4096} & x \\ 
        
        \texttt{Llama2 13b$^{2}$}& \href{https://huggingface.co/mtc/meta-llama-Llama-2-13b-hf-arxiv-summarization-5000-last_merged}{Llama-2-13b-arxiv-4096} & \href{https://huggingface.co/mtc/meta-llama-Llama-2-13b-hf-pubmed-summarization-5000-last_merged}{Llama-2-13b-pubmed-4096} & x \\
        
        \texttt{Mistral 7b$^{2}$} & \href{https://huggingface.co/mtc/mistralai-Mistral-7B-v0.1-arxiv-summarization-5000-last_merged}{Mistral-7B--arxiv-4096} & \href{https://huggingface.co/mtc/mistralai-Mistral-7B-v0.1-pubmed-summarization-5000-last_merged}{Mistral-7B-pubmed-4096} & \href{https://huggingface.co/mtc/mistralai-Mistral-7B-v0.1-govreport-summarization-1000-last_merged}{Mistral-7B-govreport-4096} \\
        \bottomrule
    \end{tabular}
    
    \caption{Links to all fine-tuned models repositories. The value `x' implies that the model was not evaluated under those settings. $^{1}$/$^{2}$/$^{3}$ indicate fine-tuning with 1k, 5k, and 10k instances, respectively.}
    \label{tab:finetuned-model-links}
    \end{table*}

\paragraph{Language Models} We used HuggingFace Transformers~\citep{wolf-etal-2020-transformers} and Microsoft Deepspeed library for distributed training.\footnote{\url{https://github.com/microsoft/DeepSpeed}} We fine-tuned BART\footnote{\url{https://huggingface.co/facebook/bart-base}} and PEGASUS-X\footnote{\url{https://huggingface.co/google/pegasus-x-large}} on the training split and a context window of 1024 and 4096, respectively. All models were fine-tuned for 4 epochs with a learning rate of $8e-4$ and batch size of 64.

\paragraph{Large Language Models} We included  Llama2 (7b)\footnote{\url{https://huggingface.co/meta-llama/Llama-2-7b}},  Llama2 (13b)\footnote{\url{https://huggingface.co/meta-llama/Llama-2-13b}}, and Mistral AI \footnote{\url{https://huggingface.co/mistralai/Mistral-7B-v0.1}} for LLM fine-tuning. We fine-tuned the models for 1 epoch using the HuggingFace Trainer API and LoRA on a training subset consisting of samples with a maximum length of 4096, such that they can fit in the context window without truncation. Since \citet{Zhou2023LIMALI} argue that 1k samples are enough to fine-tune LLMs, we experimented with 1k, 5k, and 10k training samples. Since models do not show any performance increase when trained on more than 5k samples, we opted to train on Llama2 (13b) and Mistral AI on 5k samples. We selected the LoRA parameters r=64, alpha=16, and a dropout of 0.1. Furthermore, we used the paged AdamW optimizer with a beta2 value of 0.999 and a learning rate of $2e-4$ with a constant learning rate strategy. We did not fine-tune Vicuna, since we only used the non-instruction tuned models in this setting. We excluded Falcon from fine-tuning as it only supports a context window of 2048, and therefore, it cannot be fairly compared against the other models with a context window of 4096.

\section{LLM Prompting}
\label{apx:prompt}
 Table~\ref{tab:prompt} and Table~\ref{tab:da-prompt} illustrate the prompts used to generate summaries and to score the domain adaptation of summaries using GPT-4, respectively.  For evaluation, we use the prompts introduced by ~\citet{liu-etal-2023-g} for Coherence and Fluency. However, we craft our own prompt that asseses model's ability to adapt to a new domain by evaluating the generated summaries.

\begin{table*}
\small
\centering
\begin{tabular}{p{15.4cm}l}
\textbf{0-SHOT PROMPT}\\
\toprule 

You are an expert at summarization.
Proceed to summarize the following text.\\
TEXT: \{article\}\\
SUMMARY:
\newline
\\
\textbf{FEW-SHOT PROMPT}\\
\toprule 

You are an expert at summarization. Proceed to summarize the following text.\\
TEXT: \{article\}\\
SUMMARY: \{summary\}\\
Proceed to summarize the following text.\\
TEXT: \{article\}\\
SUMMARY: \{summary\}\\
… \\
TEXT: \{article\}\\
SUMMARY:
\\

\end{tabular}
\caption{The prompt in the Benchmark for generation of domain-specific summaries using Large Language Models.}
\label{tab:prompt}
\end{table*}

\begin{table*}
\small
\centering
\begin{tabular}{p{15.4cm}l}
\textbf{SYSTEM PROMPT}\\
\toprule 
You will be given one summary written for an article. Your task is to rate the summary on one metric. Please make sure you read and understand these instructions very carefully. Please keep this document open while reviewing, and refer to it as needed.\\
 Evaluation Criteria: \{criteria\}\\
 Evaluation Steps: \{steps\} \\
\\
\textbf{USER PROMPT}\\
\toprule 
Example:\\
 Source Text: \{document\}\\
 Summary: \{summary\}\\
 Evaluation Form: Please provide your response in two parts. First the evaluation score for the defined metrics, followed by an explanation for the score. Please limit your response to 30 words.\\

\{metric\_name\} \\
 \\
\textbf{DOMAIN ADAPTATION SCORE CRITERIA}\\
\toprule 
Domain Adaptation(1-5) - the degree to which the summary adheres to the domain-specific language. A good summary employs domain-specific terminology and conveys the sense that model comprehends and encapsulates domain-specific knowledge. It resembles the content that would authored by a domain expert. Annotators were also asked to penalize summaries that didn't   adhere to domain-specific knowledge, and rather used simple words.\\
\\
\textbf{ DOMAIN ADAPTATION SCORE STEPS}\\
\toprule 
 1. Read the article carefully and understand the domain it belongs to. \newline
 2. Read the summary and check if it contains domain-specific terminologies and concepts, and if it is able to concisely summaries the domain specific concept in the article. \newline
 3. Assign a score for domain adaptation based on the Evaluation Criteria. \\
\end{tabular}
\caption{The Domain Adaptation (DA) prompt in the Benchmark to evaluate the domain adaptation capabilities of the models using a GPT-4.}
\label{tab:da-prompt}
\end{table*}

\section{Sample Summaries}
\label{apx:sample-output}
 Table~\ref{tab:sample-summaries} shows the summaries generated by Llama2 7b under zero-shot, two-shot and fine-tuning setting.

\begin{table*}
\small
\centering
\begin{tabular}{p{15.4cm}l}
\textbf{Zero-shot Llama2 7b}\\
\toprule 
mirnas are small non \- coding rnas that play important roles on the onset and progression of cancer . distinct mirnas regulate various cellular processes including cell cycle regulation , apoptosis , migration and invasion . altered expression of mirnas has been observed in breast cancer and is believed to play important roles including cell proliferation regulation , cell survival and tumorigenesis . furthermore , mirnas have been shown to play a critical role in cancer stem cell maintenance and tumorigenesis . emerging evidence suggests that mirnas may also be a attractive therapeutic target for cancer . understanding the specific functionalities of distinct mirnas and their role in disease may provide a new approach for cancer diagnosis and potentially new therapeutic strategies. reference listing: lee , j.a. (2011). mi \- rna : a duel \- role player in breast cancer progression . breast cancer : a global view . springer . 10.1007\/978\-3\-642-13570\-4\_19 . 19 . 277 \- 286 . abstract: micro \- rnas
\newline
\\
\textbf{Two-shot Llama2 7b}\\
\toprule 

Small non-coding RNAs, such as siRNA, miRNA, piRNA, and snRNA, are transcribed into mRNA but remain untranslated in eukaryotic cells. They play a crucial role in regulating gene expression and have been implicated in the onset and progression of tumorigenesis. Emerging evidence suggests that small non-coding RNAs may represent a novel cancer therapy. Unlike mRNA, small non-coding RNAs are transcribed but never translated. Some small non-coding RNAs are transcribed from non-coding regions between genes, while others are transcribed together with coding mRNAs from the coding region of the genome. Altered expression of small non-coding RNAs has been observed in breast cancer, resulting in aberrant gene expression. Studies have shown that a high frequency (~72.8\%) of gene copy number abnormality in small non-coding RNA-containing regions occurs in human breast cancer. Decreased expression of certain small non-coding RNAs, such as mir-17/20, has been
\newline
\\
\textbf{Fine-tuned Llama2 7b}\\
\toprule

 mirnas are small non - coding rnas that regulate gene expression and tumorigenesis . 
 mirnas are involved in the regulation of the cell cycle , cellular proliferation , metastasis and drug resistance of breast cancer . 
 the expression of mirnas is dysregulated in breast cancer . 
 the dysregulated mirnas may serve as novel biomarkers for breast cancer . 
 mirnas may serve as targets for gene therapy either alone or as an adjuvant to conventional therapy .

\end{tabular}
\caption{Summaries generated by Llama2 7b under zero-shot, two-shot and fine-tuning setting for a sample article (id = 2) from PubMed test set.}
\label{tab:sample-summaries}
\end{table*}

\end{document}